# Implementation of Hand Detection based Techniques for Human Computer Interaction

Amiraj Dhawan, Vipul Honrao
Dept of Computer Engineering
Fr. C. Rodrigues Institute of Technology, Vashi
Navi Mumbai, India

## ABSTRACT
The computer industry is developing at a fast pace. With this development almost all of the fields under computers have advanced in the past couple of decades. But the same technology is being used for human computer interaction that was used in 1970's. Even today the same type of keyboard and mouse is used for interacting with computer systems. With the recent boom in the mobile segment touchscreens have become popular for interaction with cell phones. But these touchscreens are rarely used on traditional systems. This paper tries to introduce methods for human computer interaction using the user's hand which can be used both on traditional computer platforms as well as cell phones. The methods explain how the user's detected hand can be used as input for applications and also explain applications that can take advantage of this type of interaction mechanism [1] [2].

## 1. INTRODUCTION
In today's age a lot of research is done for finding effective techniques and methods to make existing systems more reliable and efficient. One of the most important parameter to make system efficient and reliable is Human Computer Interaction (HCI). Many systems provide simple techniques for HCI, most common techniques for input to the systems include use of mouse, keyboard etc. These are physically in contact with the system. Recently new techniques have been developed to make interaction with the system more efficient. This paper presentsthree interactive techniques for Human Computer Interaction using the hand detection. The paper provides efficient and reliable hand based interaction techniques, which can be implemented using specific methodologies according to user's applicationrequirement. The techniques described in the paper requires preprocessing like transformation of the users hand image from one form to another, enhancements etc for implementation.These preprocessing tasks can be done using various methods depending on the environment around in which the system is being used by the specific user. This paper tries to compare the different methods for all the preprocessing tasks required and then create a pipeline of these preprocessing tasks and then use the detected hand as an interaction device for HCI applications. Next part contains the three different types of interaction with machines explained under Extraction of Input in implementation. These three different types of interaction are: Finger Counting, Hand Orientation and Finger Tracking. Further the paper explains a few applications where these different interaction mechanisms are used to enhance the interaction between the user and machine.

## 2. TECHNOLOGIES USED
### 2.1 OpenCV
OpenCV (Open Source Computer Vision Library) is a library mainly aims at real-time computer vision systems, developed by Intel. The library is platform independent. It mainly focuses on real time Image Processing and Computer Vision. Computer Vision is a science and technology that is used to process images to obtain relevant information. OpenCV library is written in C language with its primary application interface available in C. It provides wrapper classes for use in C++ language as well. Also there are full interfaces available for various other languages such as Python, Java, MATLAB and Octave. It provides basic data structures for Matrix operations and Image Processing with efficient optimizations. It provides fast and efficient data structures and algorithms which can be used for complex real time Image Processing and Computer Vision applications [3].

### 2.2 C++
C++ (pronounced as see plus plus) is a multi-paradigm, general-purpose and compiled programming language. It comprises of a combination of high-level language features like OOPS concepts and low-level language features like memory access using pointers etc. It added object oriented concepts like classes to the C language. We use the C++ interface of OpenCV for the implementation. The C++ interface provides a powerful data structure Mat (short for Matrix) which is efficient, portable and easy to use. Mat is basically just a multi-channel and multi-dimensional array which can be used to store images, intermediate image transformations, values just like a normal Array.

## 3. IMPLEMENTATION
### 3.1 Hand Segmentation
Hand segmentation deals with separating the user's hand from the background in the image. This can be done using various different methods. The most important step for hand segmentation is Thresholding which is used in most of the methods described below to separate the hand from the background.

Thresholding can be used to extract an object from its background by assigning intensity values for each pixel such that eachpixel is either classified as an object pixel or a background pixel. Thresholding is done on the input image according to a threshold value. Any pixel with intensity less than the threshold value is set to 0 and any pixel with intensity more than the threshold value is set to 1. Thus the output of thresholding is a binary image with all pixels 0 belonging to the background and pixels 1 represent the hand. Therefore the white blob that is pixels having value 1 is the object area. In our case the object is the user's hand. The most important component for thresholding is the threshold value. There are various methods to select the appropriate threshold value for better result explained in the further points [5].

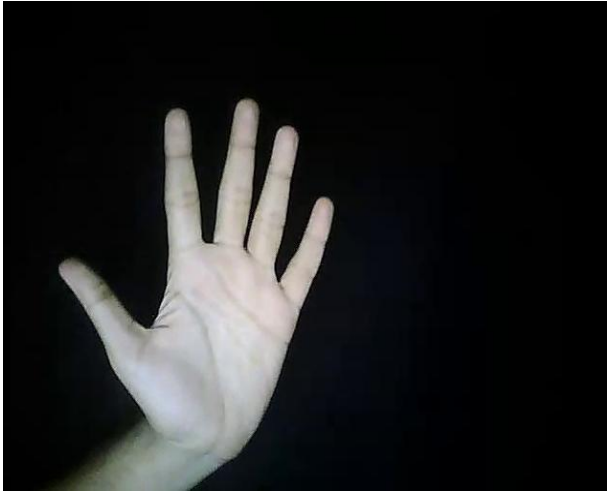

**Figure 1: Input Image Frame**

Hand Segmentation methods are basically divided into 2 broad types, one is with background constraints and the other one is with relaxed background constraints. These two types are explained as follows:

### 3.1.1 With Background Constraint
In this type of segmentation, some constraints are put on the background to extract hand blob without much noise. In this type of hand segmentation, intensity of the pixels is used for segmenting the user's hand. Typically intensity of hand is much higher, so by keeping background dark, hand can be easily segmented. This type includes the following methods for segmentation:

### 3.1.1.1 Static Threshold Value
Image frame is taken as input from the webcam in the RGB format. This image is converted into grayscale. Then either a static threshold value is used or a threshold value is selected from 0 to 255 according to the user specification which acts as the threshold value. This threshold value should be chosen by the user in such a way that the white blob of the hand is segmented with minimum noise possible. A Trackbar can be provided to adjust the threshold value for the current usage scenario.

For every usage, either the thresholdingvalue is static that is each time same value is used or the user is required to set the threshold value to ensure good level of hand segmentation. Thus this method is not used since it puts the systems success or failure dependant on the user setting a proper threshold value or on the quality of the static threshold value.This method is useful where the intensity of the hand is almost similar whenever the system is used. Also the background intensity should be similar every time the system is used. But even in constant lighting conditions during every system use, the system might fail depending on the user's hand color. If the user's hand is also darker in color, the system might not be able to separate the user's hands and the dark background. The figures 2 and 3 below show the thresholdedinput image from figure 1 using a static threshold value of 70 and 20. The noise introduced in figure 3 clearly shows how using a bad thresholding value can introduce noise which can reduce the accuracy of hand detection.

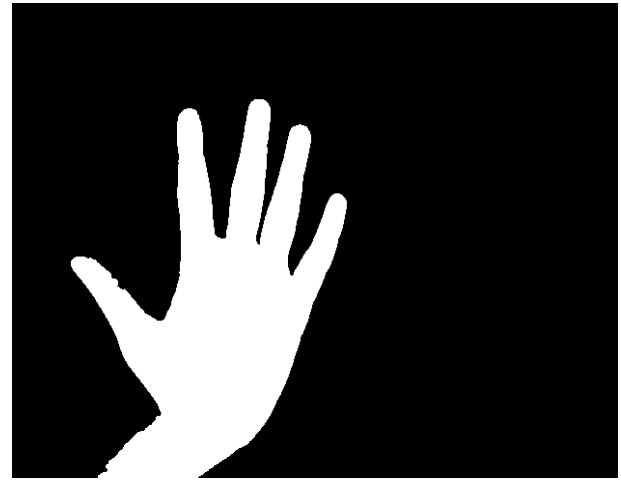

**Figure 2:Thresholded Image with threshold value as 70**

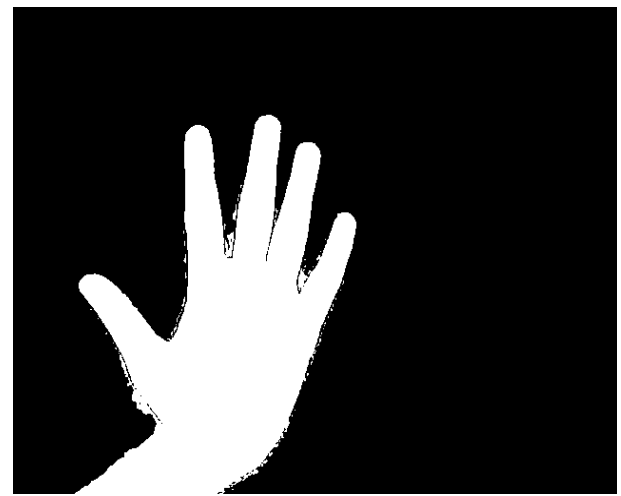

**Figure 3: Thresholded Image with threshold value as 20**

Thresholding can be performed in C++ using OpenCV using the following function [6]:

```
doublethreshold(InputArraysrc,
           OutputArraydst,
           double thresh,
           doublemaxval,
           int type);
```

- `src` - is of data type InputArray and contains the source image or frame

- `dst` - is of data type OutputArray and is the destination space for the thresholded image

- `thresh` - is of type double and it is the threshold value that is to be selected

- `maxval` - is only used in some of the different ways of applying the thresholding function. In the current case of binary thresholding does not use this variable

- `type` - defines the type of thresholding applied on the image. In our case this is THRESH_BINARY

*3.1.1.2 Incremental Thresholding Value*

In this method, same pre-processing as in the static thresholding value is done on the image input frame, converting from RGB to Grayscale. Instead of using a constant value for every input image frame, the threshold value is incremented till a condition is not met. For this method, a minimum threshold value is set and then the input image frame is thresholded using this value. If the current thresholding value does not fulfil the condition, then the thresholding value is incremented and again the same procedure is followed till the condition is met. The condition to detect hand is until only one white blob is present in the thresholded image. The detected white blob can also be some other object so whether the detected object is a hand or not is decided by the hand detection part explained further.

This method can automatically select the threshold value. This method generally gives good results especially in the environment where intensity values of input image frame changes continuously. This method ensures that the entire hand is detected as a whole blob without any internal fragmentation. But on the negative side, sometimes the background pixels near the hand might also get included in the white blob. Also if the background is not constantly dark, some areas of the background might also add up in with the hand in the white blob at certain threshold values and still make only one white blob. That is even though it would pass the condition that only white blob is present but the white blob would consist of the hand and the lighter background areas that are connected to the hand.

To remove these problems a test has to be conducted to find if the white blob has a structure similar to hand or not using convexity defects explained further.

In this method, the function described in 3.1.1.1 is used i.e. Static Thresholding. But in this case the thresholding value keeps on incrementing, that is the variable thresh is incremented by value '1' in a range from 20 to 160, till we detect only one contour in the input image.

*3.1.1.3 Thresholding using Otsu's Method*

Otsu's Methodis used to automatically select a threshold value based on the shape of the histogram of the image. This algorithm assumes that the image contains two dominant peaks of pixel intensities in the histogram that is two classes of pixels. The histogram should be bi-model for using this method. The two classes are foreground and background. The algorithm tries to find the optimal threshold value using which the two classes are separated in such a way that their intra-class variance or combined spread is minimal. The threshold value given by the Otsu's Method thus in our case works well since the images contain two type of pixels, background pixels and hand pixels. Thus the two classes are background and hand. So the threshold value tries to separate the peaks in order to give minimal intra-class variance. The example output is given in the figure 4.

Using OpenCV library this is done by passing a special parameter THRESH_OTSU as the variable type in the thresholding function given in 3.1.1.1 Static Thresholding. The thresholding function ignores thevalue in the variable thresh which was used to pass the threshold value in earlier cases. But in place of that the function calculates the threshold value using the Otsu's Method and threshold the image as per this threshold value which minimizes the within group variance.

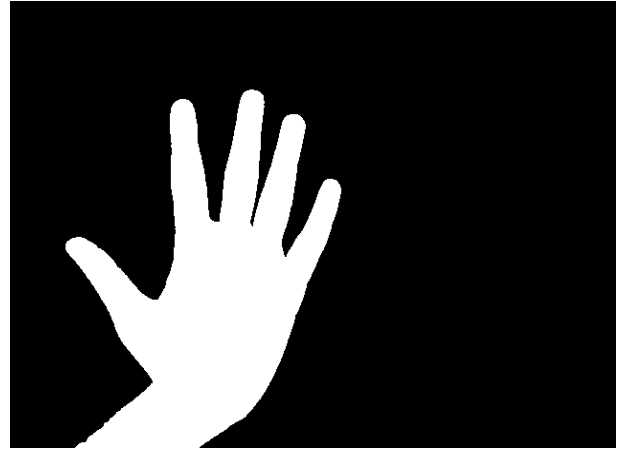

**Figure 4: Thresholded Image using Otsu Thresholding**

The advantage is that this method works well under any circumstances until the hand and background pixels create distinct peaks in the histogram of the image. The only problem with this method is that if the user's hand is not in view, the method would give a threshold value which breaks up the background pixels into two separate classes making it difficult to understand that the hand is not in view. This problem can be solved again by using the tests explained further to make sure the detected white blob is a hand only. Since the chances of the background getting thresholded in a way that the white blob passes the hand detection test are extremely less, the system practically does not give any false positives.

*3.1.1.4 Dynamic Thresholding using Color at Real Time*

Unlike previous methods of thresholding, in this method color based thresholding is done. This can also be termed as color level slicing. Initially the user has to give some dummy input image frames with the hand to be detected in the central part of the image. The system would do the analysis on these dummy input frames and generating dynamic threshold values in RGB. In this analysis, a small central circular part, with arbitrary radius, of the dummy input frames is considered initially. The first two pixels of the central part are set as minimum pixel value and maximum pixel values. Then rest all pixels in the central part are processed. For every pixel value that is scanned, it is compared with the minimum and maximum pixel values. If the scanned pixel value is less than the minimum pixels value then the minimum pixel value is updated to the scanned pixel value. Similarly if scanned pixel value is more than that of maximum pixel value, then the maximum pixel value is updated to the scanned pixel value. The range defined by the Minimum and Maximum pixel value is used to threshold the image, whichever pixel comes between this ranges is considered as hand pixel.

This method is very accurate to segment the hand if the intensity of the hand does not alter much during usage. It can detect any color of hand thus making it independent of the user's skin color. The dummy input frames should have the hand in the central part else the entire system collapses since the range decided is not actually for the hand. The background should not contain pixels with values that fall in between the decided range as they too would be included as hand pixels.

### 3.1.2 Relaxed Background Constraints

#### 3.1.2.1 Color based Thresholding

In color based thresholding, static values of hand color are considered for thresholding. RGB values of hand are taken with minimum RGB value and maximum RGB value as a range. These ranges are selected after analysing the general range of color of human hands. Then the input image frame uses these two minimum and maximum RGB values for thresholding. Any pixel between this range is considered as the hand pixel so it is set to 1 and pixels outside this range are considered as background pixel is set to 0. As there are no background constraints in this method, it is highly prone to noise.

This method can also work on general backgrounds with a slight constraint that the background should not contain pixels that lie between the ranges specified. Else extra processing like selecting the largest contour explained further is required to ensure such white blobs are not detected as hand. The thresholding values are very tricky to select. For some user's,the color of the hand could vary a lot and be outside the specified range thus making the system unable to detect such user's hand.

#### 3.1.2.2 Background Subtraction

This method is useful when the background remains static throughout usage. There are no special constraints on the background except there should not be any or much change in it. An image of the background is provided to the system without the user's hand in view. Using this background image provided, the process of background subtraction is done. Whenever a new image frame is processed, it is converted into gray scale. Then the gray level background image is subtracted from the gray level of the image frame. The difference between the background image and the current image frame is the hand which is in view of the image frame but not in the background image. Thus the user's hand is easily segmented in any type of background.

This system works very well with any kind of background. The hand is very accurately segmented irrespective of the color of the hand or the backgroundcolor. The only negative with this system is that most of the times it is impossible to have a completely static background. If the background changes much in the current image frames, the difference between the current background and the initial background can also become significant enough to cause the system to give false positives.

## 3.2 Hand Detection

### 3.2.1 Contours

A contour is the curve for a two variables function along which the function has a constant value. A contour joins points above a given level and of equal elevation. A contour map illustratesthe contour using contour lines, which shows the steepness of slopes and valleys and hills. The function's gradient is always perpendicular to the contour lines. When the lines are close together, the magnitude of the gradient is usually very large. Contours are straight lines or curves describing the intersection of one or more horizontal planes with a real or hypothetical surface with.

The contour is drawn around the white blob of the hand that is found out by thresholding the input image. There can be possibilities that more than one blob will be formed in the image due to noise in the background. So the contours are drawn on such smaller white blobs too. Considering all blobs formed due to noise are small, thus the large contour is considered for further processing specifying it is contour of hand.

In this implementation, after preprocessing of the image frame, white blob is formed. Contour is drawn around this whiteblob. Vector contains set of contour points in the coordinate form. Figure 5 shows the detected contour for the input image.

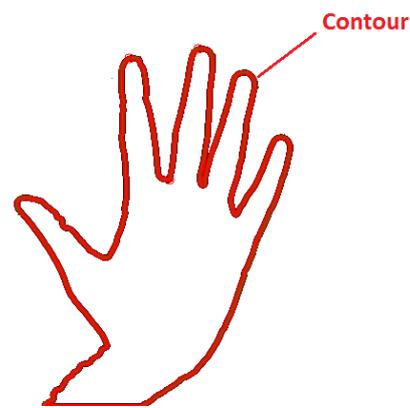

**Figure 5:Detected Contour for the Input Image**

For finding contours in OpenCV using C++ as the language the following function is used [6]:

```
voidfindContours(
     InputOutputArray image,
     OutputArrayOfArrays contours,
     int mode,
     int method,
     Point offset=Point());
```

- `image`– is the source image

- `contours`– is a vector of vector of Point data structure defined by OpenCV

- `mode` – is the contour retrieval mode. We use CV_RETR_EXTERNAL which retrieves only the extreme outer contours. Other modes are also available like CV_RETR_TREE, CV_RETR_CCOMP, CV_RETR_LIST

- `method`– is to select the contour approximation method. We use CV_CHAIN_APPROX_SIMPLE which is capable of compressing vertical, horizontal and diagonal segments and stores only their end points. This provides sufficient accuracy with lesser storage requirement.

Other methods are CV_CHAIN_APPROX_NONE, CV_CHAIN_APPROX_TC89_L1, CV_CHAIN_APPROX_TC89_KCOS

- `offset` – this is an optional parameter to add an offset to each point stored for the contour

To draw the detected contours on a new white colored image, we use the following function provided in OpenCV for C++:

```
voiddrawContours(
      InputOutputArray image,
      InputArrayOfArrays contours,
      intcontourIdx,
      const Scalar&color,
      int thickness=1,
      intlineType=8,
      InputArray hierarchy=noArray(),
      intmaxLevel=INT_MAX,
      Point offset=Point() );
```

- `image` – is the input image on which the contours should be drawn. In our case this is a new white image of same size as input image

- `contours-` is a vector of vector of Point data structure defined by OpenCV. These are the contours that we want to draw

- `contourIdx` – index of the contour to be drawn from the contours vector. If this value is negative then all the contours in the parameter contours are drawn

- `color` – is of the data type Scalar and it is a constant value. This is the color to be used to draw the contours

- `thickness` – defines the thickness of the line to be used to draw the contours. If this value is negative, then the contours are filled with the color specified. Default value is set to 1

- `lineType` – is to select how the lines are drawn whether they are 8-connected line, 4-connected line and CV_AA for antialiased line

- `heirarchy` – this is optional information only required when only some of the contours are required to be drawn

- `offset` – optional offset to add to contour points while drawing

### 3.2.2 Convex Hull

The convex hull of a set of points in the euclidean space is the smallest convex set that contains all the set of given points. For example, when this set of points is a bounded subset of the plane, the convex hull can be visualized as the shape formed by a rubber band stretched around this set of points.

Convex hull is drawn around the contour of the hand, such that all contour points are within the convex hull. This makes anenvelope around the hand contour. Figure 6 shows the convex hull formed around the detected hand.

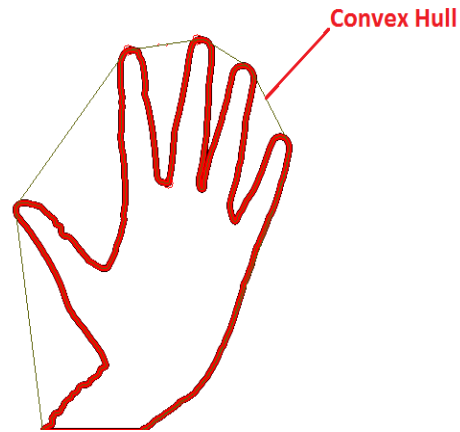

**Figure 6: Calculated ConvexHull of the Input Image**

For detecting the convex hull of a contour the following function is used [6]:

```
void convexHull(
      InputArray points,
      OutputArray hull,
      bool clockwise=false,
      boolreturnPoints=true );
```

- `points` –this is the set of points for which the convex hull is calculated. In our case we would pass the largest contour as a set of points to find the convex hull around the said contour

- `hull` – this acts as the storage for the hull output. Either it is a vector of indices of the points parameter if all the hull points are present in the points vector or the points itself if all the hull points do not exist in the points parameter

- `clockwise` – this is the orientation flag. If it is true then the output convex hull is oriented in the clockwise direction. Else in the anti-clockwise direction. By default it is set to false

- `returnPoints` – if the points parameter is a vector in place of a matrix which it is in our case, the returnPoints variable is ignored. Else in case of a matrix, this flag decided whether the hull output storage stores the indices of points or the points itself. By default this is set to true

The convex hulls are also drawn on an image using the same function drawContours as explained in section 3.2.1 Contours. This is because both, contours and convex hulls are nothing but a collection of points which needs to be connected with straight lines.

### 3.2.3 Convexity Defects

When the convex hull is drawn around the contour of the hand, it fits set of contour points of the hand within the hull. It uses minimum points to form the hull to include all contour points inside or on the hull and maintain the property of convexity. This causes the formation of defects in the convex hull with respect to the contour drawn on hand.

A defect is present wherever the contour of the object is away from the convex hull drawn around the same contour. Convexity defect gives the set of values for every defect in the form of vector. This vector contains the start and end point of the line of defect in the convex hull. These points indicate indices of the coordinate points of the contour. These points can be easily retrieved by using start and end indices of the defect formed from the contour vector. Convexity defect also includes index of the depth point in the contour and its depth value from the line. Figure 7 shows an example of convexity defects calculated in the detected hand using the input image of figure 1.

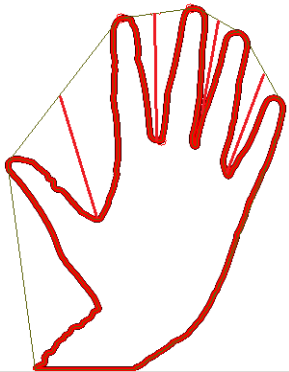

**Figure 7:Major Convexity DefectsCalculated using the Input Image of figure 1**

For finding convexity defects using the contour and its convex hull we use the following function [6]:

```
void convexityDefects(
     InputArray contour,
     InputArrayconvexhull,
     OutputArrayconvexityDefects);
```

- `contour` – contains the largest interesting contour points

- `convexhull` – contains the calculated convex hull of the contour passed

- `convexityDefects` – acts as the storage for the output convexity defects. This storage is a vector of convexity defects. Every convexity defect is represented by a 4 element integer vector that is Vec4i defined by OpenCV. These four integer points are start_index, end_index, farthest_pt_index and fixpt_depth. The start_index and end_index are the indices of points in the contour parameter which are the start and end point of a particular convexity defect. The farthest_pt_index is also an index of a point in the contour which does not lie on the convex hull and is the farthest distance from it for a particular convexity defect. fixpt_depth is the approximation of the distance between the farthest point and the hull.

The `fixpt_depth` is the parameter which we use to remove convexity defects which are not large enough to be considered as the space between two fingers. Due to this requirement on the minimum depth required for consideration as the space between two fingers, the system may not detect some fingers if the user's hand is far away from the web camera.

## 3.3 Extraction of Input

After all pre-processing on the input image frame mentioned in hand segmentation and hand detection is done, useful information is extracted from the user's hand for input purposes. For extracting useful input, three main things that are mentioned in hand detection. Contour and convexity defects in the image frame are mainly used for extraction of inputs. For interaction with the computer system, various commands can be given

In this paper we propose three techniques for interactions as follows:

### 3.3.1 Finger Counting

In this method, the count of fingers from the user's hand is extracted. It makes use of convexity defects for detecting the finger count. As mentioned above, the data structure of convexity defect gives the depth of the defects. Many such defects occur in a general hand image due to wrist position and orientation. But some defects have far greater depth than others. These are the gaps between the two fingers. As shown in the figure 8, count given by the user is two. There are many defects that are formed, but the depth of defect formed due to the gap between two fingers is much greater and thus can be separated from other non-important defects. For two adjacent fingers, there is one such defect.

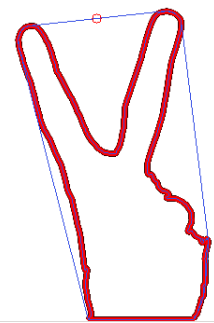

**Figure 8: Finger Count of 2**

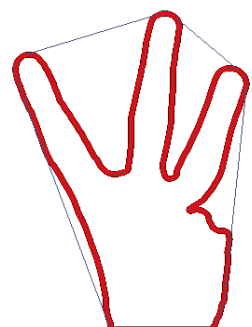

**Figure 9: Finger Count of 3**

Similarly in figure 9, count given by the user is three that is having two large defects in the image. So finger count is easily determined by adding one to the count of large defects in the images. This technique fails when there is no large defect in the input image frame. This situation occurs when there is one finger or no finger in the image. In both the cases there is no large defect, so it cannot be determined whether no count that is count 0 is given by the user or count one is given by the user. To overcome this issue we design the commands from count of 2 to 5. The count of 1 and 0 can also be used as a command but they both should represent the same command.

### 3.3.2 Hand Orientation

In this method of interaction, the orientation of hand is taken into consideration. Four possible orientations can be recognized such as up, down, right and left. Same as previous method the data structure of convexity defects are used to determine the orientation of hand. In the vector of convexity defect the defect with the maximum depth is found out. This gives the index of starting and ending points of line of the largest defect in the contour. These two point's coordinate can be easily extracted from the contour vector using the index values. Using these two points, the midpoint of the line is calculated, using

$$mid\_point = \frac{(start\_point + end\_point)}{2}$$

This midpoint is compared with depth point of the defect, and orientation of hand is determined.

Calculated midpoint and depth point are compared in two ways by checking initial conditions. For this the difference between x-coordinate and y-coordinate of midpoint and depth point is used. If there is small difference in x-coordinate and large difference in y-coordinate then the orientation of hand is up or down. For detecting whether it is up or down, difference between y-coordinates is used. As origin of the image is at top left, so if the difference between y-coordinates of the midpoint and depth point is negative then orientation of hand is up. And if it is positive then orientation is down. As shown in the figure 10, difference between y-coordinate of the midpoint and depth point is positive, so the orientation of hand is down.

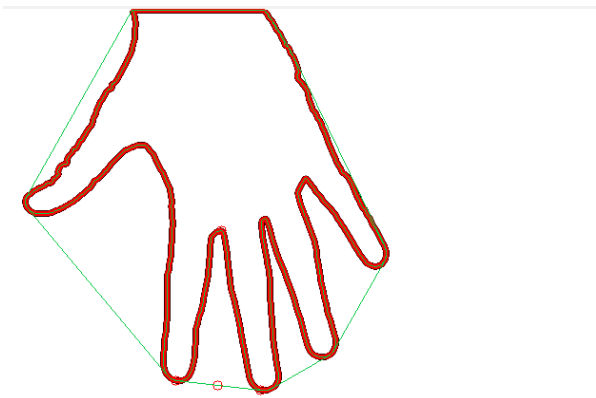

**Figure 10: Down Orientation**

Similarly in figure 11, the line of hull with defect having maximum depth is considered. The difference between x-coordinates of the midpoint and depth point is much larger than the difference between y-coordinate. So orientation of the hand is either right or left. If the difference between midpoint and depth point is positive then orientation of hand is towards right. And if the difference between midpoint and depth point is negative then orientation of hand is towards left.

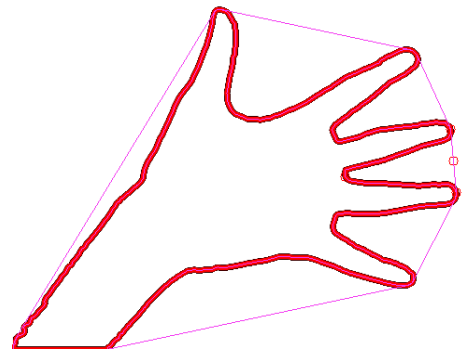

**Figure 11: Right Orientation**

### 3.3.3 Finger Tracking

In this method the fingertip of the user is used as the input pointer. The input pointer is just like a mouse pointer. First the detected hand contour is tested for structural properties which include the no of convexity defects and the approximate depth of these defects in proportion to the hand contour size. Then the top most point of the contour is used as the mouse pointer. This top most point is the fingertip of the user. To emulate a click, the user is required to keep the fingertip stable with slight in acceptable error range. The user interface should have large clickable areas so that the number of miss-clicks is less. The acceptable error gives a circular area around the detected fingertip in which the fingertip should be for a fixed number of frames in order to register it as a click. Sometimes the fingertip is not detected in consecutive frames due to various reasons so a relaxation is provided of 3 frames. So if the fingertip is inside the acceptable error area for a fixed number of frames with the relaxation of those 3 consecutive frames may not have the fingertip in the acceptable area.

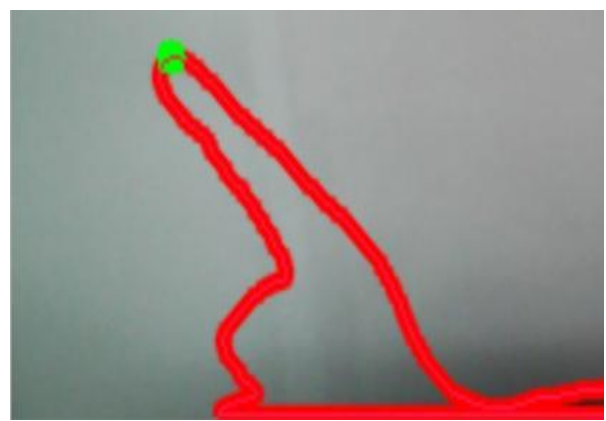

**Figure 12: User's tracked hand. The green tip acts as the pointer**

Every time the fingertip goes outside the current acceptable area, a counter is set to 0. The new location in the current frame becomes the center of the acceptable area. Now for each frame if the fingertip is detected in the acceptable area then the counter is incremented. If the fingertip is not detected in the acceptable area then a different variable anti-counter is incremented. Then if anti-counter reaches the value of 2 then counter is reset and the current location is made the center of the acceptable area. If the fingertip is detected in the acceptable area then anti-counter is reset and counter is incremented. If the counter reaches a fixed value then the click is registered and the counter reset. Anti-counter variable is used to implement the relaxation of 2 consecutive frames not having fingertip in the acceptable area due to any reason.

## 4. APPLICATIONS
### 4.1 Gesture Controlled Robot using Image Processing[8]

Service robotics involves directly interacting with people through various traditional interaction techniques like Remote Control etc. Now a days there is a need to find a more natural and easy to use interaction system through which systems can be made user friendly. The implementation of various Human Computer Interaction described in this paper can be used for controlling robots through gesture for giving commands. Out of 3 methods of HCI specified above two of them can be used for controlling robots through gesture as follow.

One method that can be used is Orientation of Hand. As there can be four possible orientations of hand that can be detected in the provided system, each of these orientations can be used to specify the direction of movement to the robot to move in particular direction and navigate in the environment. Another method that can be used is Finger Count. Each Finger Count can be associated with a command for the robot's movement. By changing the finger count, the direction of the robot can be changed according to it. For all five finger counts commands like Forward, Backward, Right, Left and Stop can be given to the robot.

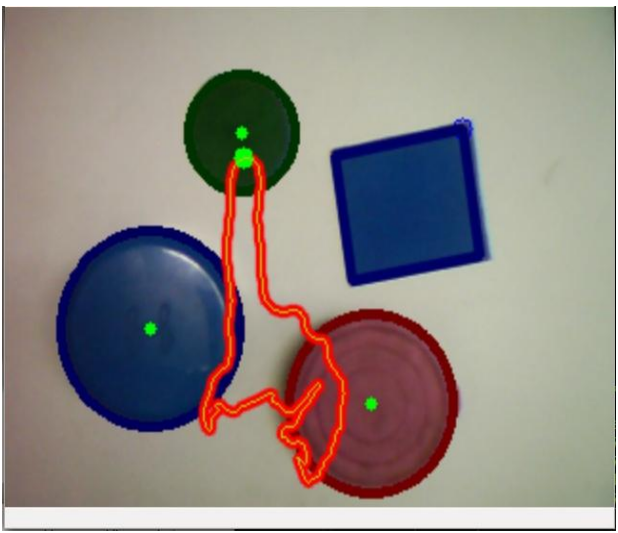

**Figure 13: Selection of Object using the User's Hand Pointer for Pick and Place Robot**

### 4.2 Pick and Place Robot [9]

A pick and place robot is basically a robotic arm that is used to pick up object and place them at a destination. This application uses finger tracking for interaction between the user and the system. The camera on top of the robotic arm can remotely send the work envelope of the robot to the user at a remote location. At the remote location the user is supposed to select the object which should be picked up and then also select the destination for this object. This can be done by showing the fingertip overlay over the images coming from the robot's camera. When a click is registered inside a detected object, the object is marked as the target object and the arm picks up the object and then again the destination is the point where the next click is registered from the finger tracking module. The object is thus placed at the destination.

## 5. FUTURE WORK
The current system does not have a good background subtraction method which can be used in any situation. Thus the current system puts lots of constraints on the user for successful working. The future work includes reducing these constraints so that the system is usable in more scenarios. Also the detection can be improved to be more accurate and produce better and precise results.

## 6. CONCLUSION
In the current system we have implemented various techniques for efficient human computer interaction. We have also provided different techniques for preprocessing of input images. These provided techniques can be used accurately and efficiently applying some constraints.

## 7. ACKNOWLEDGMENTS

We would like to thank Mr. H.K. Kaura, HOD of computers department Fr.C.R.I.T and Prof. Krutika the department coordinatorfor the their support and guidance provided during the work on the implementation of these techniques and while writing this paper.